\documentclass{ieeeaccess}
\usepackage{graphicx}
\usepackage{subfig, cite, booktabs, multirow, tikz}
\usepackage{amsmath}
\usepackage{siunitx}
\usepackage{textcomp}
\usepackage{amssymb,amsfonts}
\usepackage{algorithmic}
\usepackage{color}
\usepackage{bm}

\NewSpotColorSpace{PANTONE}
\AddSpotColor{PANTONE} {PANTONE3015C} {PANTONE\SpotSpace 3015\SpotSpace C} {1 0.3 0 0.2}
\SetPageColorSpace{PANTONE}%

\def\BibTeX{{\rm B\kern-.05em{\sc i\kern-.025em b}\kern-.08em
    T\kern-.1667em\lower.7ex\hbox{E}\kern-.125emX}}

\usetikzlibrary{shapes, arrows}
    
\begin{document}

\doi{undefined yet}

\title{ Learning-Based Human Segmentation and Velocity Estimation Using Automatic Labeled LiDAR Sequence for Training }

\author{\uppercase{Wonjik Kim}\authorrefmark{1,2},
\IEEEmembership{Member, IEEE},
\uppercase{Masayuki Tanaka}\authorrefmark{1,2},
\IEEEmembership{Member, IEEE}, \uppercase{Masatoshi Okutomi}\authorrefmark{1},
\IEEEmembership{Member, IEEE}, and \uppercase{Yoko Sasaki}.\authorrefmark{2},
\IEEEmembership{Member, IEEE}}

\address[1]{Department of Systems and Control Engineering, School of Engineering, Tokyo Institute of Technology, Meguro-ku, Tokyo 152-8550, Japan (e-mail: wkim@ok.sc.e.titech.ac.jp; mtanaka@sc.e.titech.ac.jp; mxo@sc.e.titech.ac.jp)}
\address[2]{Artificial Intelligence Research Center, National Institute of Advanced Industrial Science and Technology, Koto-ku, Tokyo 135-0064, Japan (e-mail:y-sasaki@aist.go.jp)}

\markboth
{Kim \headeretal: Learning-Based Human Segmentation and Velocity Estimation Using Automatic Labeled LiDAR Sequence for Training}
{Kim \headeretal: Learning-Based Human Segmentation and Velocity Estimation Using Automatic Labeled LiDAR Sequence for Training}

\titlepgskip=-15pt

\corresp{Corresponding author: Wonjik Kim (e-mail: wkim@ok.sc.e.titech.ac.jp).}

\begin{abstract}
In this paper, we propose an automatic labeled sequential data generation pipeline for human segmentation and velocity estimation with point clouds. Considering the impact of deep neural networks, state-of-the-art network architectures have been proposed for human recognition using point clouds captured by Light Detection and Ranging (LiDAR). However, one disadvantage is that legacy datasets may only cover the image domain without providing important label information and this limitation has disturbed the progress of research to date. Therefore, we develop an automatic labeled sequential data generation pipeline, in which we can control any parameter or data generation environment with pixel-wise and per-frame ground truth segmentation and pixel-wise velocity information for human recognition. Our approach uses a precise human model and reproduces a precise motion to generate realistic artificial data. We present more than 7K video sequences which consist of 32 frames generated by the proposed pipeline. With the proposed sequence generator, we confirm that human segmentation performance is improved when using the video domain compared to when using the image domain. We also evaluate our data by comparing with data generated under different conditions. In addition, we estimate pedestrian velocity with LiDAR by only utilizing data generated by the proposed pipeline. 

\end{abstract}


\maketitle

\section{Introduction}
\label{Introduction}
\PARstart{R}{obot} navigation depends on real-time, precise, and robust sensing. A robot should recognize its surrounding environment and objects such as pedestrians and other robots. The robot is also required to be robust for various kinds of situations. Therefore, LiDAR is often employed to acquire accurate 3D information with a high sampling frequency. For example, LiDAR has been utilized for mapping processes~\cite{schwarz2010lidar} and robotics applications including robot navigation~\cite{niijima2018autonomous}. Human recognition with LiDAR is a very important task in robot navigation. In many LiDAR systems, several sensors scan to acquire 3D information and the scanning frequency is very high. Therefore, a single scanned data is considered the same as frame data similarly to how video data are collected for processing. This is positive because we can use the advantages of high performance computer vision algorithms.

In the computer vision field, videos have been researched with various approaches including action recognition~\cite{jiang2014thumos}, video retrieval~\cite{Li_2015_ICCV}, and irregular detection~\cite{boiman2007detecting}. Because deep neural networks have had a positive impact on computer vision, especially in the image domain~\cite{Jonathan, Farabet, DBLP}, neural networks for the video domain are now being actively researched~\cite{SlowFast,Tran_2015_ICCV,Gu_2018_CVPR}. Another topic in this field of research is semantic video object segmentation. This focuses on the detection and segmentation of object-like areas in video with predefined class labels. Since deep neural networks have achieved high performance in the image segmentation field, various network architectures have also been proposed for handling the video domain~\cite{Hong_2017_CVPR,Seguin_2016_CVPR}. However, the performance of the learning based approach strongly depends on the training dataset. Available video datasets are not labeled at pixel-level with the entire frame. Therefore, many learning based semantic video segmentation methods take weakly supervised learning to overcome the lack of ground truth ~\cite{Seguin_2016_CVPR, Hong_2017_CVPR}. On the other hand, in this study, we tackled data collection by generating sequential data. We also focused on human segmentation with sequential data collected by LiDAR. In this paper, we define the 'frame' as data from a single LiDAR scan, and the 'sequence' as sequential data from constant LiDAR scanning.

Human recognition with LiDAR data has been researched vigorously~\cite{DBLP, wkim, kim2019automatic}. According to~\cite{kim2019automatic}, the human recognition performance drastically decreases as distance between a human and LiDAR increases. Because the number of points is inversely proportional to the square of the distance between a human and LiDAR, humans in the distance are difficult to recognize with only shape information detected from the frame. On the other hand, the sequence provides not only shape but also motion information, the recognition accuracy of distant humans can be improved by utilizing this motion information. We have confirmed that the sequence-based approach improves the accuracy of human detection when compared to the frame-based approach.

Collecting a sufficient amount of labeled data requires significant investment in terms of both time and money. In this study, we develop an automatic 3D LiDAR data sequence generation pipeline for human detection and velocity estimation. In comparison to single frame data generation, data sequence generation is more challenging because we need to take into account the temporal consistency of the sequence. For example, all objects should be temporally continuous. In addition, the generated data sequence should be spatially and temporally realistic to ensure the network is accurately trained.

In~\cite{wkim}, a method to generate a single frame of realistic 3D LiDAR data for the human detection has been proposed. In that method, a statistical human shape model~\cite{mochimaru2006dhaiba} is used to build different types of precise human shapes. That model is very useful for generating the single frame of LiDAR data. However, this information is insufficient for generating the realistic LiDAR data sequence. For sequence generation, it is necessary to consider 1) the human walking model, 2) human trajectory, and 3) the sensor position and pose. In this study, we incorporated a walking model and an observed walking trajectory. Combining this information based on the LiDAR trajectory consideration, the 3D LiDAR data sequence is generated using human labels.

One of our goals is to accelerate the research of learning-based video human segmentation with point cloud data. For that purpose, we have generated a large amount of labeled sequential data (more than 10K sequences) with the proposed data generation pipeline. We have trained several neural networks with different training policies. The trained networks have been evaluated in comparison with actual data collected by a real LiDAR sensor. All generated data and labeled real data are presented in following url. Trained network weight and test sample code also included.

http://www.ok.sc.e.titech.ac.jp/res/LHD/

The remainder of this paper is organized as follows. We quickly review related work in Section~\ref{Related work}. In Section~\ref{Proposed Data Generation Pipeline}, each step in the data generation procedure is explained in detail. The network architecture for the training sequence is described in~\ref{Network architecture for sequence training}. The specific policy of the training and experimental results are then discussed in Section~\ref{Evaluation}. Finally, we conclude our research with an outline of further improvements and potential future applications in Section~\ref{Conclusion}. 

\begin{figure*}[t]
	\begin{centering}
		\includegraphics[width = 175mm, height = 56mm]{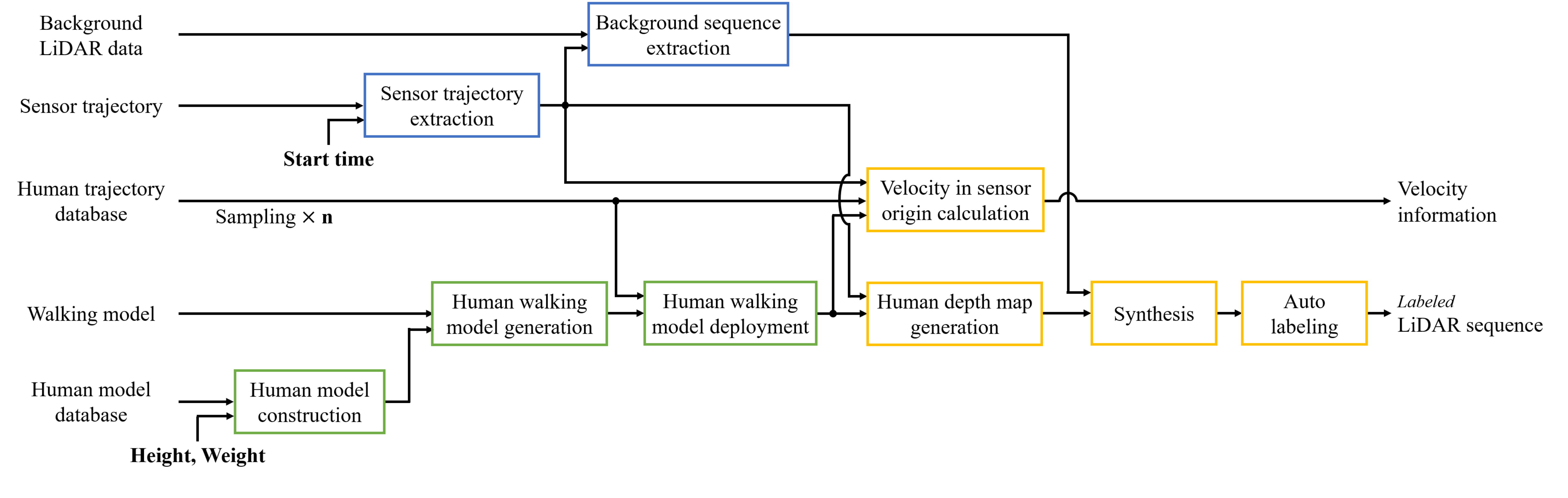}
		\caption{{\bf Overview of our data generation pipeline}. Bold words represent controllable parameters. (Green boxes: Human walking model) Human models are built by weight and height, then they are combined with a human walking motion. Thereafter, they are deployed following the sampled trajectory. (Blue boxes: Background sequence) Sensor trajectory and background LiDAR data are refined with a frame length of sequence generation. (Yellow boxes: LiDAR data generation) Human walking models are synthesized with depth map sequentially. The synthesized pixel-wise depth maps are labeled based on the information from the human model deployed location. Then, pixel-wise velocity information of sensor origin is generated according to human velocity, velocity command, and labeled information. }
		\label{fig:overviewofthepipeline}
	\end{centering}
\end{figure*}

\section{Related Work}
\label{Related work}
\noindent
{\bf Dataset of depth map.} After the release of the Microsoft Kinect in 2010~\cite{Kinect}, several RGB-D datasets have been published. RGB-D datasets for human recognition have also been provided, such as that for the re-identification of a person with RGB-D sensors~\cite{barbosa2012re}, BIWI RGBD-ID dataset~\cite{munaro20143d}, and UPCV Gait dataset~\cite{kastaniotis2015framework}. As the Kinect cannot measure depths greater than 10 [m], LiDAR sensors were employed to handle depths over 10 [m]. In addition, LiDAR sensors are used in auto driving technology. In this field, the KITTI dataset~\cite{Geiger2012CVPR} is widely used by many researchers~\cite{premebida2014pedestrian, levi2015stixelnet}. However, KITTI only provided 93K+ of depth data without labeling. Collecting labeled depth maps is still challenging. Pixel or point-wise labeling for 3D depth data is usually a challenging task that involves significant costs. Under the circumstances, the video game Grand Theft Auto was deployed to collect data ~\cite{DBLP, johnson2017driving, richter2016playing}. This approach may reduce the cost of data construction, but limitations still remain. Grand Theft Auto is not designed for research purposes; therefore, we cannot control specific properties of the circumstances of the simulation, such as the human body type and model deployment location. Automatic Labeled LiDAR Data~\cite{wkim, kim2019automatic} have been released with 1M+ data including depth, xyz coordinates and pixel-wise human labels. Automatic Labeled LiDAR Data may cover demands in the image domain; however, its application is not sufficient to address the requirements of the video domain. In addition, the datasets of depth map cannot handle velocity information because they only contain single frames, and not sequences on the time axis. 

\noindent
{\bf Datasets for RGB video segmentation.} Following the increase in video dataset demands, several datasets have been published. For example, the Freiburg-Berkeley Motion Segmentation dataset~\cite{brox2010object} has been proposed for motion segmentation. In addition, SegTrack v1~\cite{Tsai2010MotionCT} and v2~\cite{li2013video} have also been published for video segmentation specialized with fast motion and complex changing shapes. In the case of video object segmentation, the DAVIS challenge~\cite{caelles20182018} has been presented and updated since 2016. Accordingly, RGB video datasets have been vigorously proposed and updated. On the other hand, to the best of our knowledge, fully labeled video segmentation datasets for LiDAR with large scale have not been proposed. Our research focuses on filling this gap.

To address these problems, we constructed a sequential data generation pipeline enabling us to change any parameters and environments. Further, the proposed generator creates human labels and velocity information in the process, thereby incorporating this normally human related task into the computation cost. As a result, as long as sufficient computational resources are available, we can generate sequential data continuously.
\begin{figure}[t]
	\begin{center}
	\subfloat[Example 01]{
		\includegraphics[width = 35mm, height = 50mm]{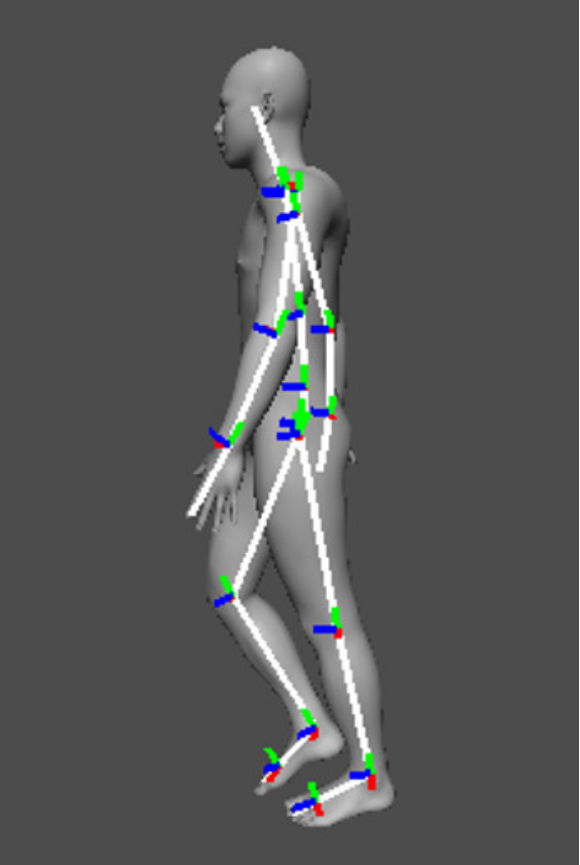}
	}
	\subfloat[Example 02]{
		\includegraphics[width = 35mm, height = 50mm]{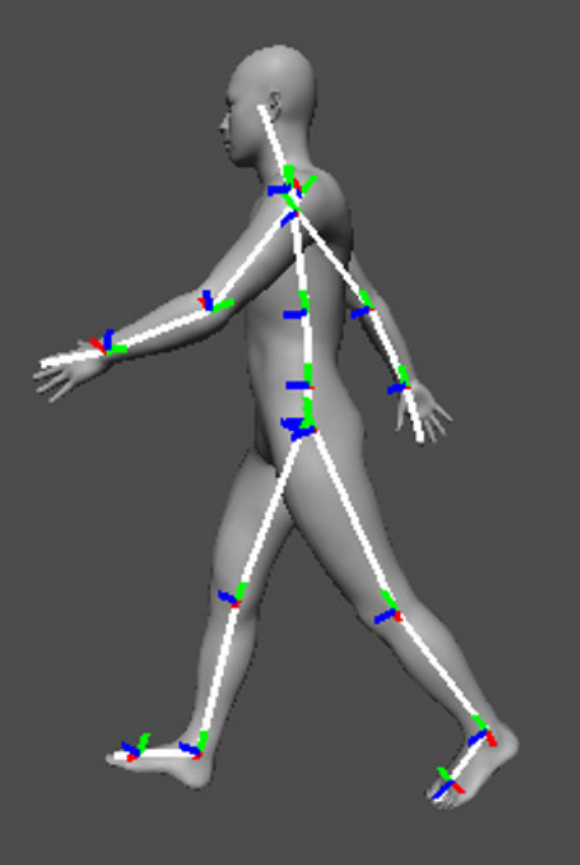}
	}\caption{ {\bfseries Examples of human models built by DhaibaWorks}. Figure (b) is at the moment of maximum distance between the left and right heels. }
        \label{exampleofhumanmesh}
	\end{center}
\end{figure}

\section{Sequential Data Generation Pipeline}
\label{Proposed Data Generation Pipeline}
\noindent
One of the main contributions of this paper is the automatic generation of labeled sequences of LiDAR data considering the precise human model and motion, without involving manual labeling. Unlike image generation, all of the object data must be connected in time-series for generating a sequence. The sequence generation pipeline comprises three steps: 1) background sequence collection, 2) human walking model generation, and 3) sequential LiDAR data generation. The details are described in the following subsections, and an overview of the pipeline is illustrated in Fig.~\ref{fig:overviewofthepipeline}.

\subsection{Sensor trajectory and background sequence extraction}
\label{subsec:Background}
\noindent
From the given sensor trajectory and background sequence, we extract a specified length of data. First, the start time is randomly selected. Then, we cut out the sensor trajectory and associated background sequence from that start time with the specified time length. Those data are used in the sequential LiDAR data generation process described in Section~\ref{subsec:LiDAR data generation}. The extracted sensor trajectory is also used for the velocity calculation and the human depth map generation process detailed in Section~\ref{subsec:LiDAR data generation}. The standardized sensor trajectory is also utilized for velocity calculation and the human depth map generation process detailed in Section~\ref{subsec:LiDAR data generation}. We can employ any LiDAR data as background LiDAR data in simulations~\cite{APOLLO, DBLP}, in real time~\cite{Geiger2013IJRR, Cordts2016Cityscapes}, and taken by ourselves. The sensor trajectory is determined automatically when the LiDAR data are determined.

\subsection{Human walking model}
\label{subsec:Human model}

\begin{figure}[t]
	\begin{center}
    \includegraphics[width = 80mm, height = 35mm]{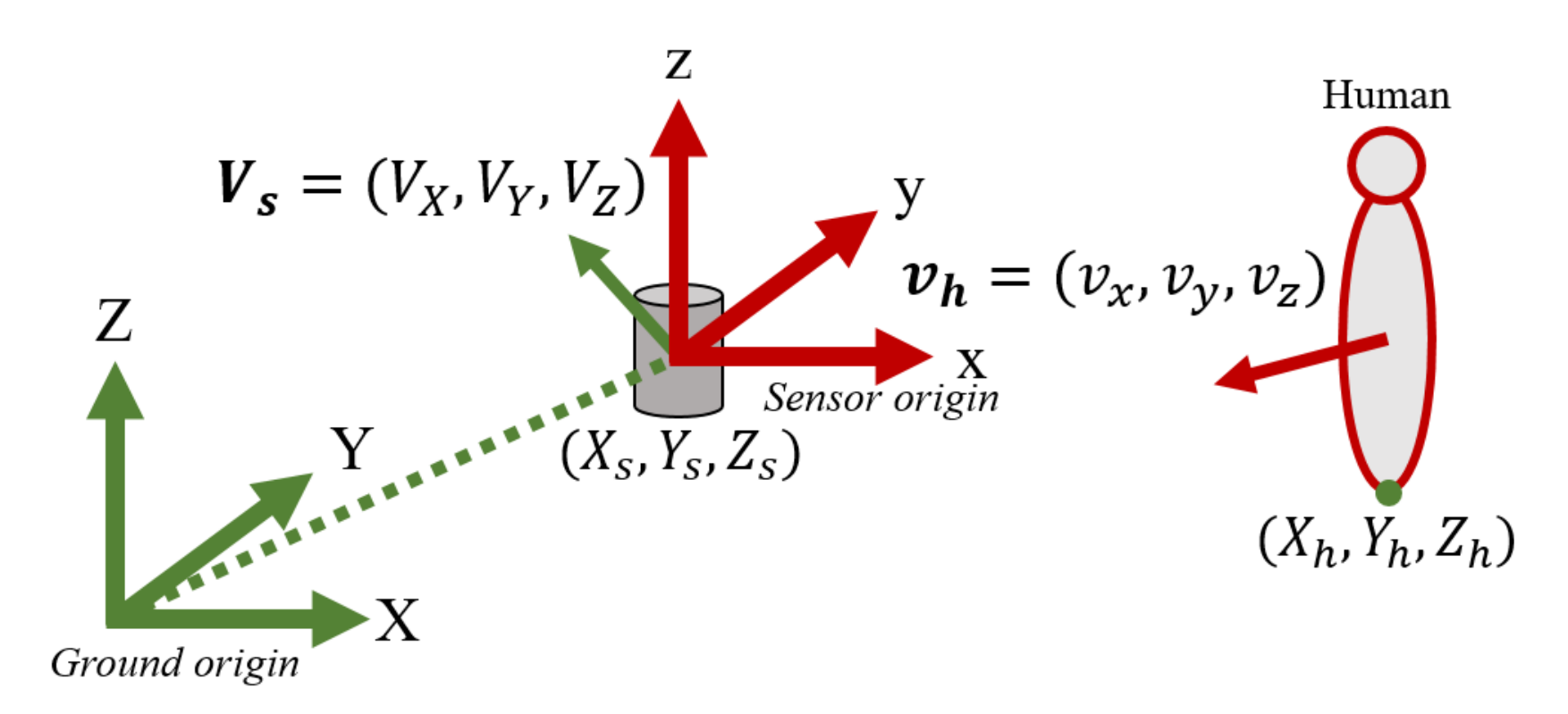} \\
	\caption{ {\bfseries Coordinate systems}. An uppercase character indicates that its origin is the ground and a lowercase character indicates that its origin is the sensor. }
        \label{coordinate}
	\end{center}
\end{figure}

\begin{figure*}[t]
	\begin{center}
	\subfloat[Point cloud of frame \#01/32]{
		\includegraphics[width = 55mm, height = 27mm]{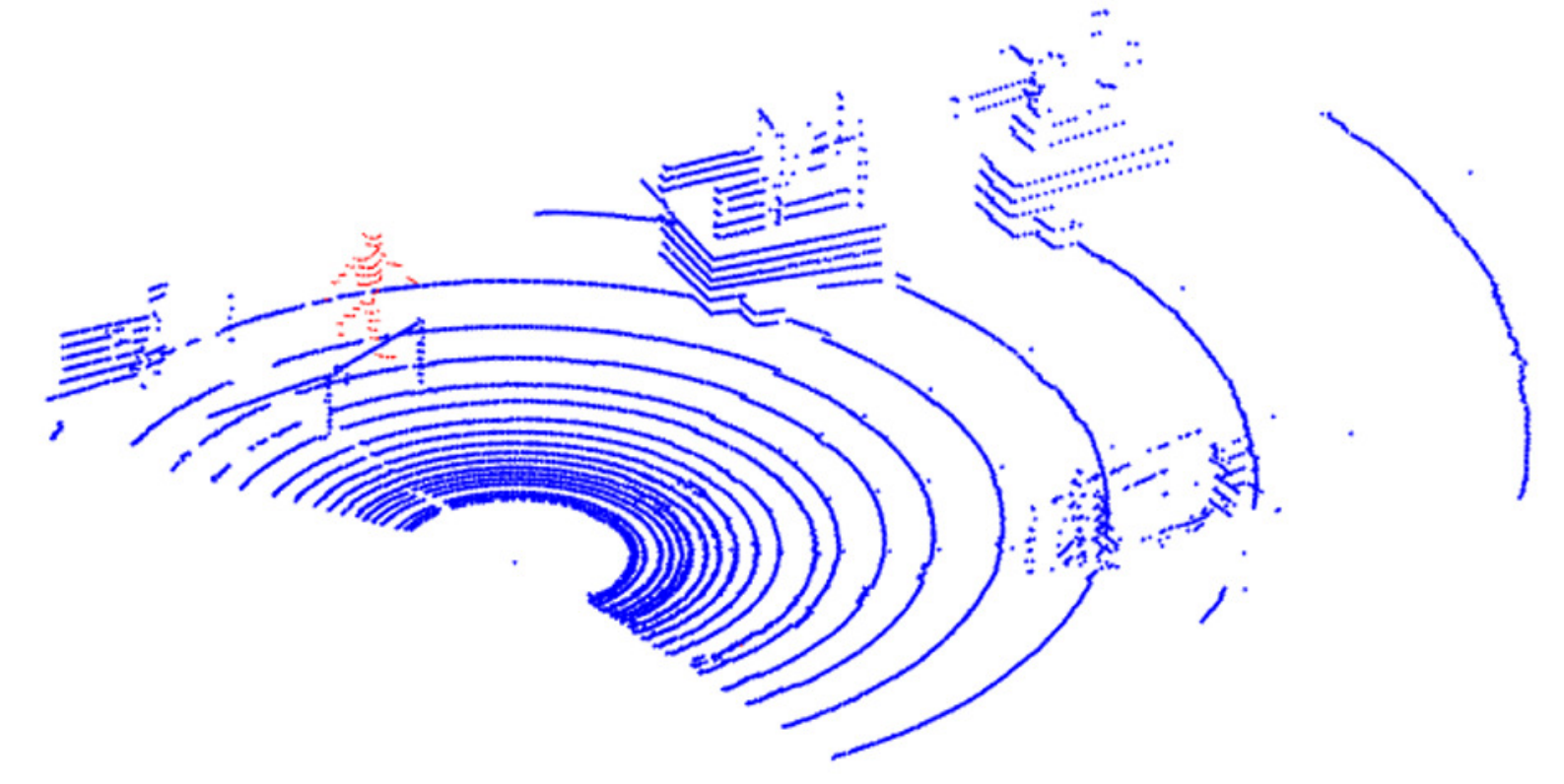}
	} 
	\subfloat[Point cloud of frame \#16/32]{
		\includegraphics[width = 55mm, height = 27mm]{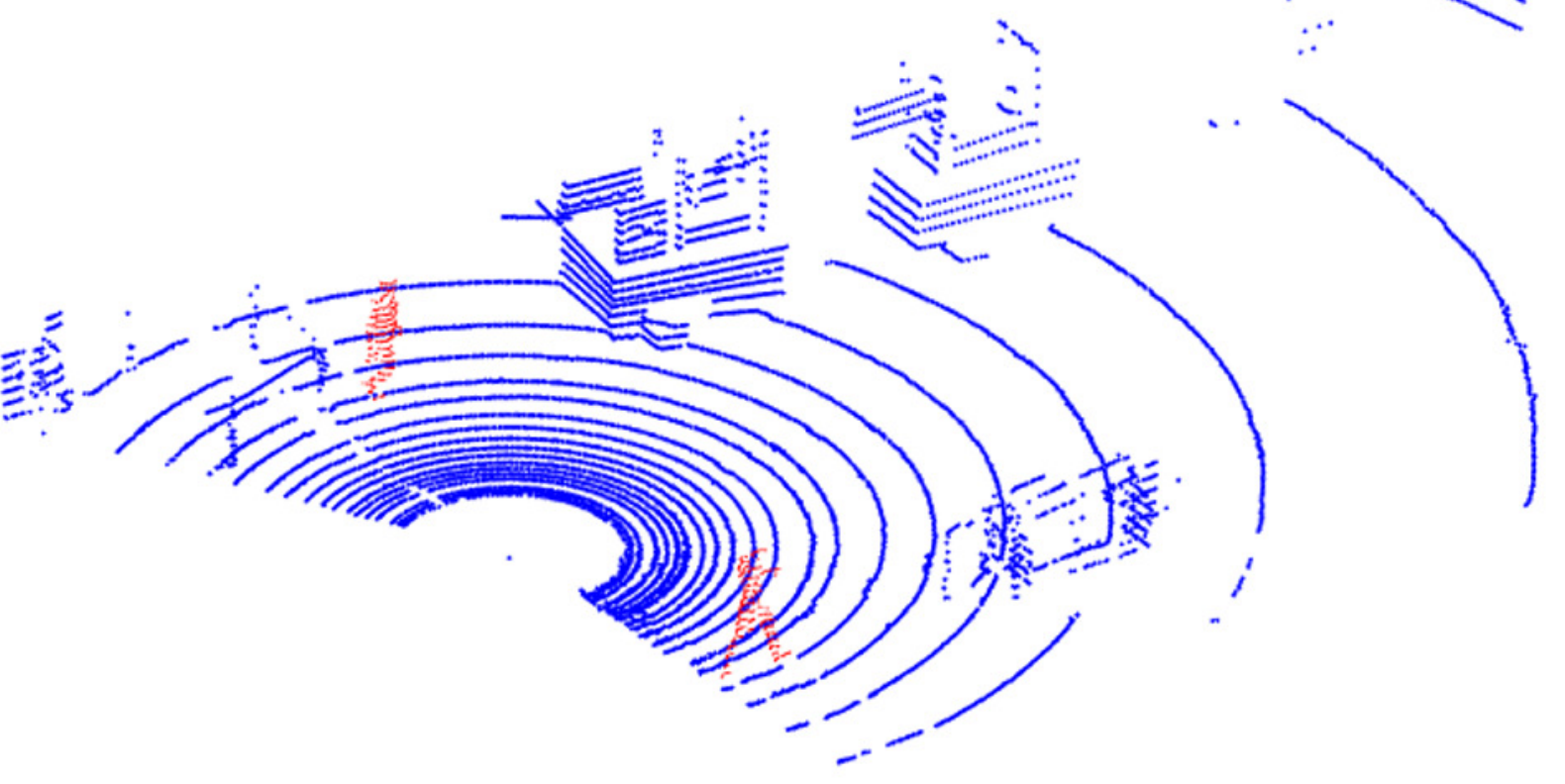}
	} 
	\subfloat[Point cloud of frame \#32/32]{
		\includegraphics[width = 55mm, height = 27mm]{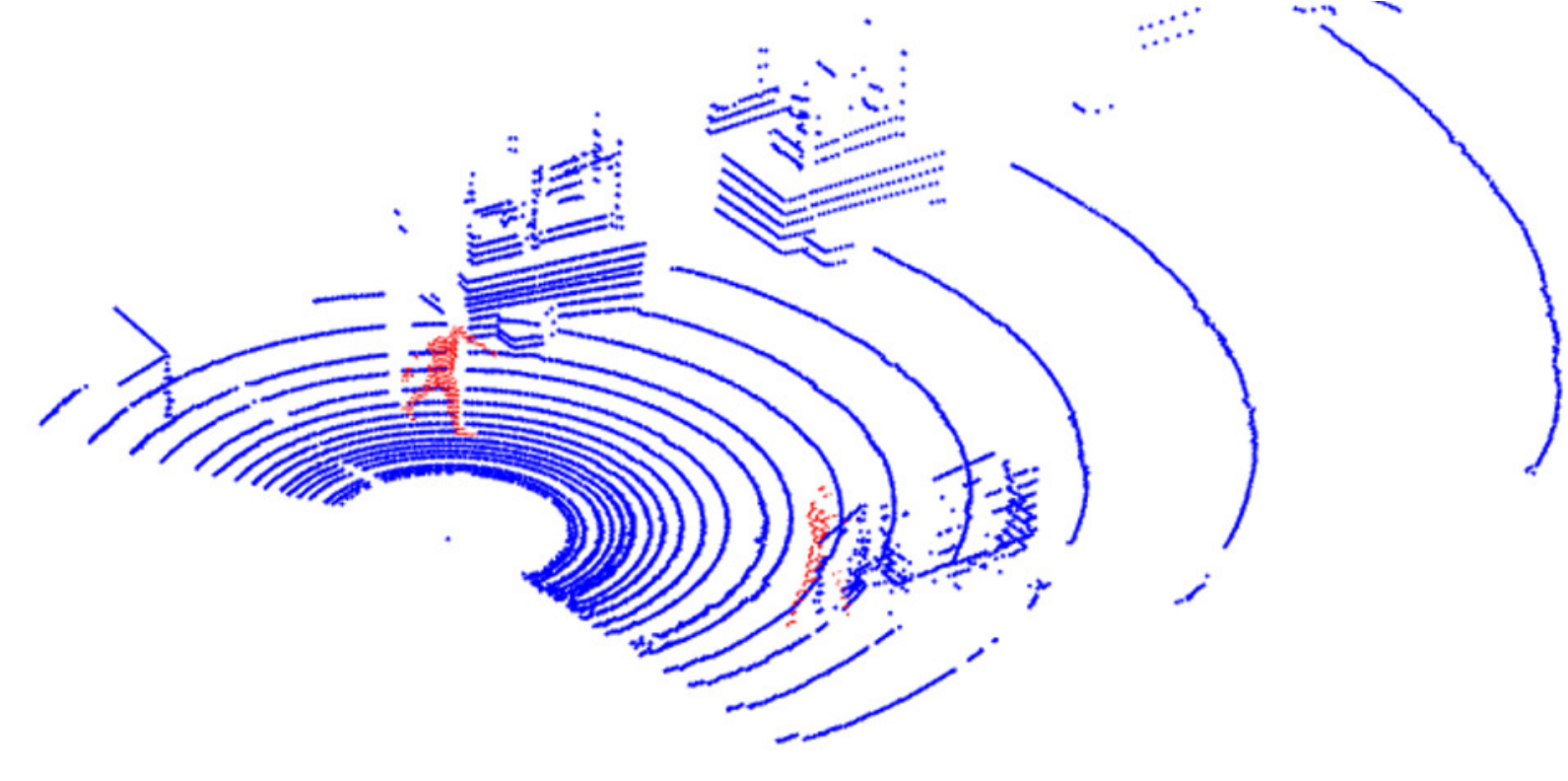}
	}
	\caption{{\bfseries Example frames of a generated sequence in point cloud}. Blue points denote the background and red points denote a human. Notice that human models are walking following their individual trajectory and the background scene is also changing following the LiDAR trajectory. }
        \label{exampleofgenerator}
	\end{center}
\end{figure*}
\noindent
For the human walking model, first, we build a human model based on the human body database~\cite{mochimaru2006dhaiba}. The walking sequence is also observed. We can generate the human walking mode by combining the observed walking sequence and the human model build with height and weight. We independently observed human trajectory. Assuming that the human model, the walking model, and the human trajectory can be sampled independently, we can construct various kinds of human walking models with associated human trajectories. However, since we considered them independently, it is necessary to contemplate the relationship between human velocity and stride length in order to construct a precise human walking model.

For the human model database, we employed a 'DhaibaWorks'~\cite{mochimaru2006dhaiba} to build precise, functional 3D human models. DhaibaWorks supports editing and visualizing basic models such as 3D meshes and skeletal structures, including human models with motion~\cite{endo2014hand}. Using DhaibaWorks, we can easily generate a specific human model by setting human parameters such as height, weight, and action status~\cite{endo2015estimation}.

\begin{figure*}[t]
	\begin{centering}
		\subfloat[Entire network]{
		\includegraphics[width = 175mm, height = 43.2mm]{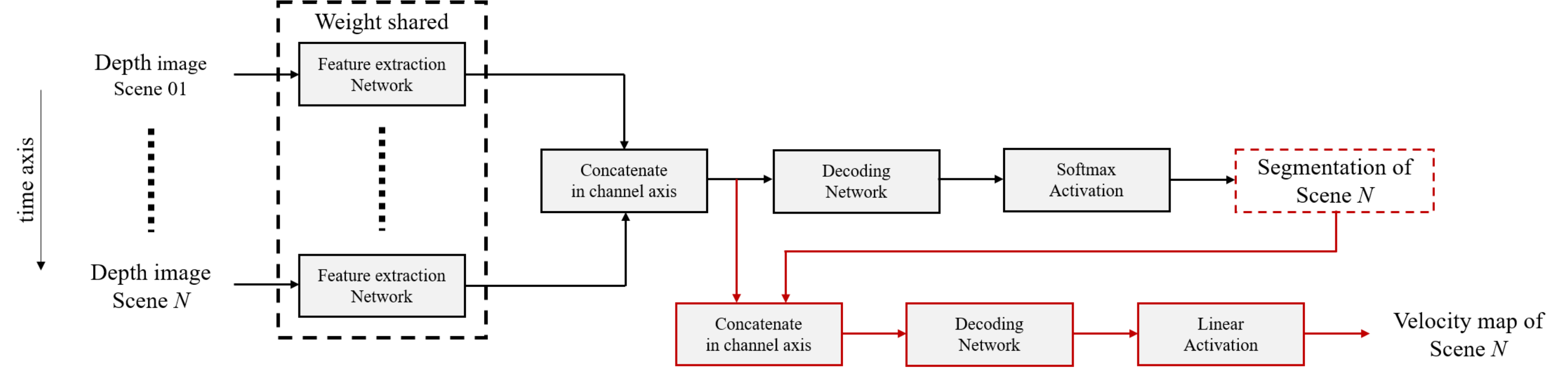}
		} \\
		\subfloat[Feature extraction network architecture]{
		\includegraphics[width = 132mm, height = 47mm]{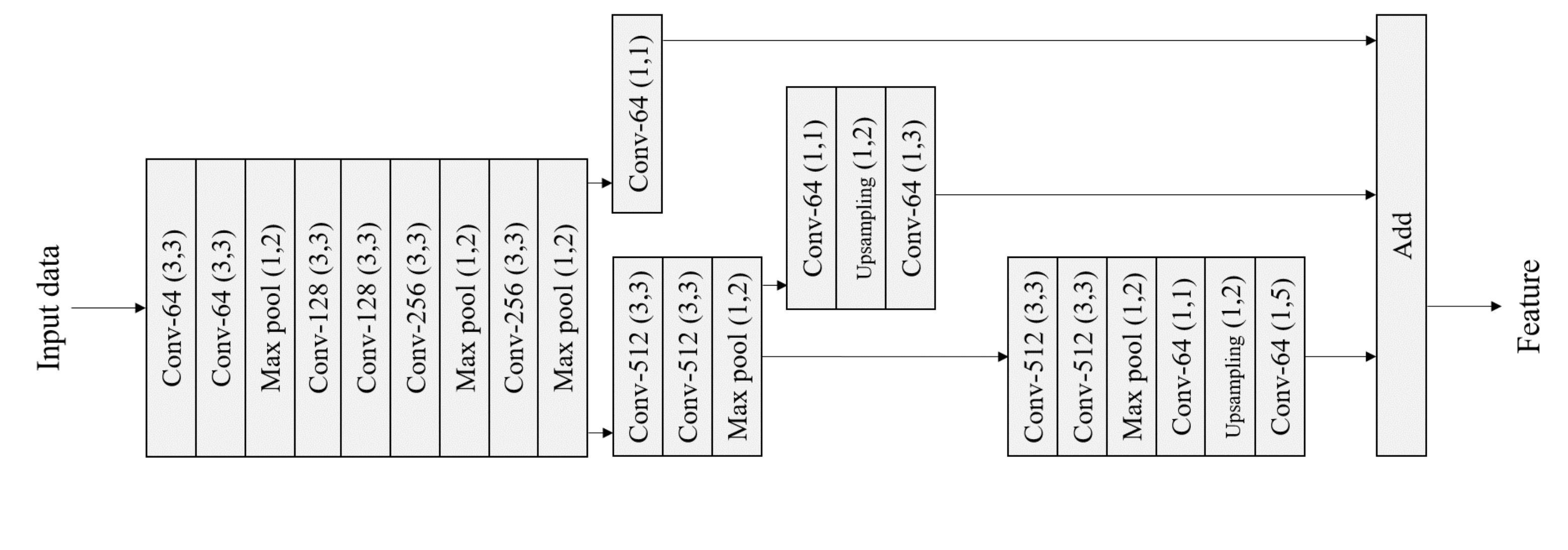}
		}
		\subfloat[Decoding network architecture]{
		\includegraphics[width = 43mm, height = 41mm]{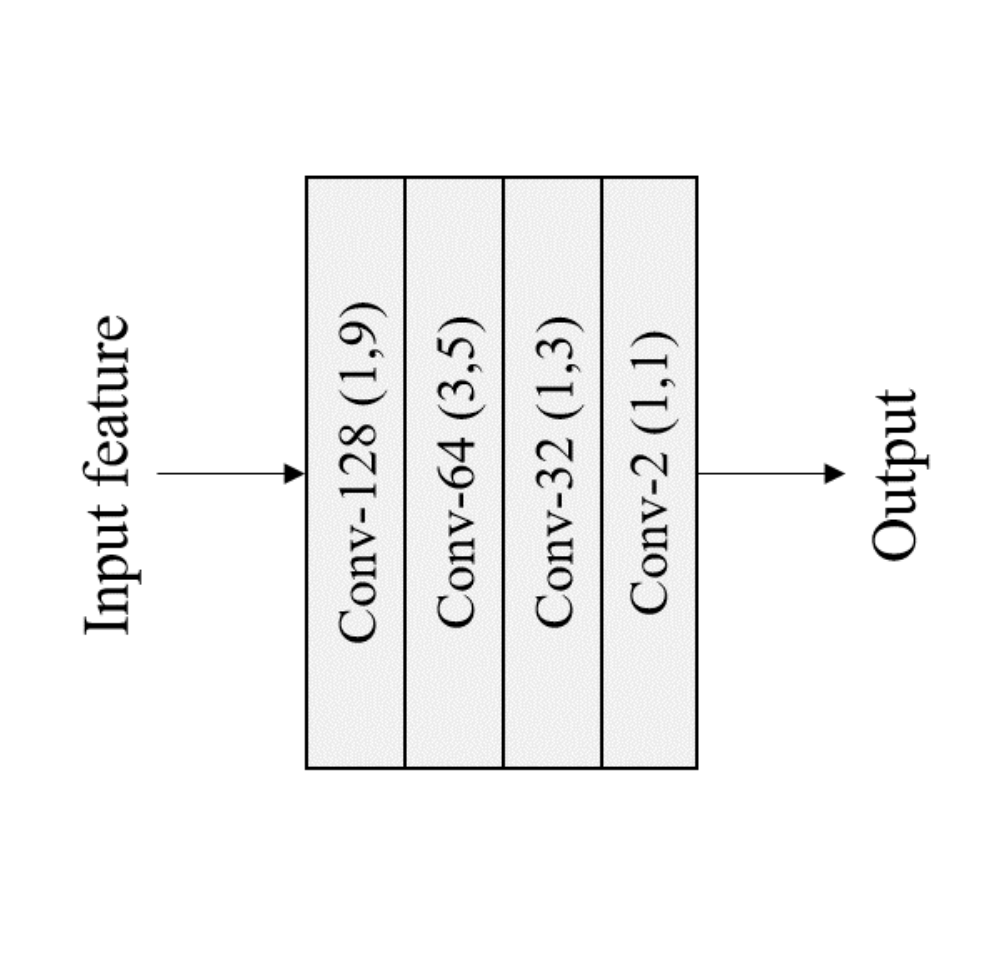}
		}
		\caption{ {\bf Network architecture for learning sequence}. This architecture can be trained with different numbers of input frames. The convolutional later parameters are denoted as "Conv-\{number of channels\} \{kernel size\}". All convolutional layers are activated with a ReLU function except "Conv-2 (1,1)" in decoding network.}
		\label{fig:learning_model}
	\end{centering}
\end{figure*}

\subsection{Sequential LiDAR data generation}
\label{subsec:LiDAR data generation}
\noindent
In the sequential LiDAR data generation step, random human models, walking trajectories, and LiDAR trajectory are sampled. Then, the depth map of the constructed human model is synthesized. Thereafter, the synthesized depth map of the human model and the background depth map associated to the LiDAR trajectory are combined to generate the training depth map for human segmentation. After generating a depth map, human models are resampled by the relationship between stride length and observed velocity in trajectory information. Thereafter, updated human models and the LiDAR position are relocated based on their trajectory information. With this loop flow, we can generate the LiDAR sequence.

The entire coordinate system in this study is illustrated in Fig.~\ref{coordinate}. To synthesize the human model depth map, we virtually located the LiDAR sensor at $(X_s,Y_s,Z_s,{\bf Q}_s)$ of ground origin, where ${\bf Q}_s$ is the sensor quaternion. Then, the human model depth map was synthesized, virtually inserting the human model at $(X_h, Y_h, 0)$ of ground origin, where $(X_h, Y_h)$ is a sampled position in the human trajectory. The human model direction is also determined by the sampled human trajectory. Once the geometrical positions of the LiDAR sensor and the human model are provided, the associated depth map can easily be synthesized. 

Once the human model depth map is synthesized, this is simply combined with the background depth map by pixel-wise minimum depth selection. The depth map taken by the LiDAR sensor usually includes holes or missing pixels whose depth could not be measured. We leave these holes as they are for the synthesis process because these types of holes are equally obtainable in a real sensing process. In addition, the human labeling task can be simultaneously performed because we know which pixels correspond to the human model depth map.

In the velocity calculation process, pixel-wise velocity information of sensor origin is generated. Human velocity of ground origin $\bm{V}_h$ , and sensor velocity of ground origin $\bm{V}_s$ can be calculated by sensor trajectory. In addition, the transform function from the ground origin to the sensor origin can also be calculated using the sensor trajectory. Therefore, the velocity information for each pixel can be generated as follows, where $p$ is one pixel in a frame:

\begin{eqnarray}
    \bm{v}_p = \begin{cases}  \bm{V}_h - \bm{V}_s \qquad  & if \; human \; pixel \\
                         -\bm{V}_s \qquad & if \; background \; pixel \\
    \end{cases}
%     \nonumber
\end{eqnarray}

Figure \ref{exampleofgenerator} shows examples of the generated sequence. Detailed information for the generating Fig. \ref{exampleofgenerator} is described in Section~\ref{DataGen}.

\section{Network architecture for sequence training}
\label{Network architecture for sequence training}
\noindent
To utilize the generated sequences for training data, we designed a network architecture for human segmentation and velocity estimation with depth images as shown in Fig~\ref{fig:learning_model}. Because convolutional layers can represent any architecture, we used a typical convolutional neural network architecture for feature extraction.

The architecture for segmentation is described in black ink in Fig.~\ref{fig:learning_model}. First we sampled depth images from the sequence. Thereafter, each input was computed by weight shared convolutional layers for producing the features. Next, we concatenated each feature in the channel axis. Then we adapted another convolutional layer to concatenated features as a decoder. For activation, we employed softmax for the segmentation task. With this architecture, we can tune the network by feeding the sampled depth sequence and the label of the last frame in the sampled sequence. We used categorical cross-entropy for the loss function of the segmentation task.

The architecture for velocity estimation is described in red ink in Fig.~\ref{fig:learning_model}. We concatenated the segmentation result and concatenated features in the channel axis. Then we adapted the other convolutional layers to re-concatenate the features as a decoder. For activation, we employed linear for the velocity estimation task. We used the Mean Square Error (MSE) for the loss function of the velocity estimation. In addition, we did not consider defected pixels in the input scene for MSE calculation.

For sequential data learning, we randomly selected one piece of sequential data from the training dataset at every step. Then, a certain frame length of sequentially sampled training data could be generated from the selected sequence.

\section{Evaluation}
\label{Evaluation}
\noindent
For evaluation, we generated 1,108 sequences as described in table~\ref{table:controlvariable}, then, 1,000 sequences were used for training and 108 sequences were used for validation and estimation. As a result, 32,000 frames were used for training. We only used 108 final frames in each sequence for estimation. We also prepared 0.1K of manually labeled real data for evaluation.

In this Section, we employed a Intersection over Union (IoU) in this experiment. A IoU was calculated as positive as human. We computed a IoU as a True Negative (TN), False Negative (FN), True Positive (TP), and False Positive (FP) as follows:

\begin{eqnarray}
    IoU & = & \frac{TP}{TP + FP + FN} %\nonumber
\end{eqnarray}

\begin{figure}[t]
	\begin{center}
    \includegraphics[width = 80mm, height = 30mm]{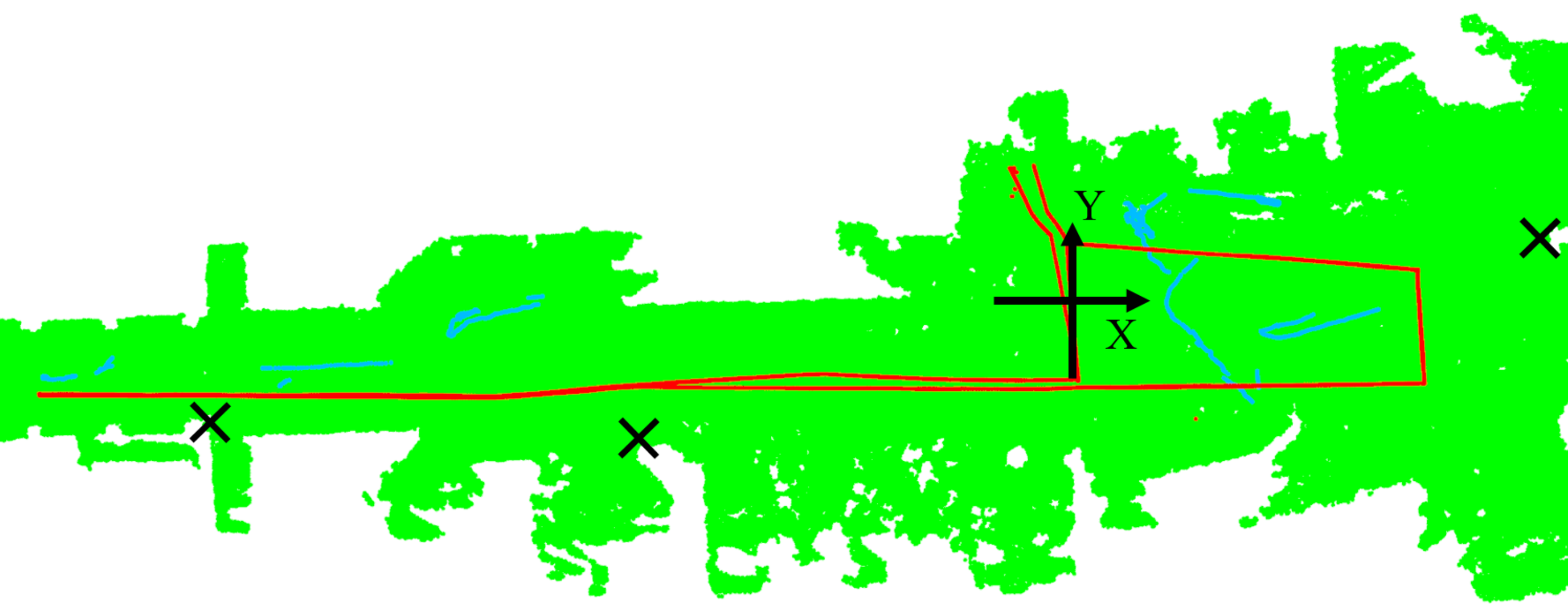}
	\\
	\caption{{\bfseries LiDAR and human trajectories}. The green points denote the floor, red points denote LiDAR trajectory, and blue points denote human trajectory. X marks denote the sensor installation position, and 10 random human trajectories are illustrated for example. The xy coordinate for ground origin is also shown. The entire length of the figure in the X-axis is approximately 50 [m].}
        \label{backgroundpointcloud}
	\end{center}
\end{figure}
\begin{table}[t]
	\begin{center}
		\caption{Data generation properties}
		{\normalsize
		\begin{tabular}{cc} \hline
			Description & Specification \\ \hline
			Number of detector pairs & 32  \\
			Limitation in horizontal scanning & 1024 \\
			Vertical scanning range & +10.67 to -30.67\textdegree  \\
			Angular resolution in vertical & 1.33\textdegree \\
			Angular resolution in horizontal	& 0.2\textdegree \\
			Human number in one frame & 1 to 30 \\
			Output velocity unit & mm/s \\
			Frame number in one sequence & 32 \\
			Data sampling rate & 10 [Hz] \\ \hline
		\end{tabular}
		}
		
		\label{table:controlvariable}
	\end{center}
\end{table}

\begin{table}[t]
	\begin{center}
		\caption{Real data for evaluation}
		{\small
		\begin{tabular}{cc} \hline
			Description & Specification \\ \hline
			Location & Miraikan~\cite{Mirai} \\ 
			Date & 2017.Aug.04 \\
			LiDAR model & Velodyne HDL32E \\
			LiDAR height from ground & $800$ [mm] \\
			Sequence number & 100 \\
			Labeling method & Manual \\ \hline
		\end{tabular}
		}
		\label{realdatatable}
	\end{center}
\end{table}

\begin{table}[t]
	\begin{center}
		\caption{Human model specification}
		
		\begin{tabular}{c|c|c|c|c|c} \hline 
			 Height & \multirow{2}{*}{1200} & \multirow{2}{*}{1400} & \multirow{2}{*}{1600} & \multirow{2}{*}{1700} & \multirow{2}{*}{1800} \\ 
			 (mm) & & & & & \\  \hline 
			 Weight & \multirow{2}{*}{15 20 30} & \multirow{2}{*}{20 30 40} & \multirow{2}{*}{40 50 70} & \multirow{2}{*}{50 60 80} & \multirow{2}{*}{50 70 90} \\
			 (kg) & & & & & \\ \hline 
		\end{tabular}
		\label{table:humanmodel}
	\end{center}
\end{table}

\subsection{Data generation setting}
\label{DataGen}
\noindent
The background sequence was collected in the Miraikan 3rd floor using a HDL-32E LiDAR. In the background sequence collection step, the velocity command to the LiDAR equipped robot and the coordinate transformation matrix from the ground origin to the LiDAR origin were also recorded. The LiDAR trajectory is estimated using 'Real-Time 6DoF Monte-Carlo Localization'~\cite{ohsato2015real}. the estimated LiDAR trajectory is shown in Fig.~\ref{backgroundpointcloud}.

In a real world environment, human beings may have multiple postures including standing, walking, and running. However, we assume that the human is always walking during the sequence generation for this study. Walking motion data is required to build an artificial human walking model. We used one period of walking data that was collected in~\cite{kobayashi2016age}. This walking motion data consists of 230 frames for a single walking motion. We estimated the walking stride length based on a distance between left and right heels. For constructing the human model, we take fifteen typical combinations of height and weight as summarized in Table~\ref{table:humanmodel}. We believe that these combinations cover a variety of relevant scenarios. For gathering human trajectory, the HOKUYO UTM-30LX sensor~\cite{Hokuyo} was installed at a fixed laser sensor position as shown in Fig~\ref{backgroundpointcloud}. With the HOKUYO UTM-30LX, we can obtain the sequential human location, direction, and velocity. Therefore, we used collected visitors' trajectories gathered on Sep 21, 22 and Dec 06, 07 in 2018~\cite{sasaki2017long}. As a result, 70,300 different kinds of walking trajectories were utilized for sequence generation. Fig.~\ref{backgroundpointcloud}shows an example of the walking trajectories collected. By observing real walking trajectories, we can also avoid deploying human models into unreachable areas. With this information, we now have 230 frames of walking motion data, fifteen combinations of the human model, and 70,300 kinds of trajectories. As such, a total of 242,535,000 different types of human walking models can be generated. 

For generating velocity information, we used the human velocity command linked to a LiDAR equipped robot, and the coordinate transformation matrix from the ground origin to the sensor origin. The human velocity is recorded as xy coordinates for ground origin. Therefore, we describe the human velocity as the xy coordinate of the sensor origin using the coordinate transformation matrix. Then, we can obtain the pixel-wise human velocity by substituting each human velocity to a whole pixel in each human label. For background velocity, we assumed that the opposite direction of velocity commands of ground origin are the same as the background velocity of sensor origin. Therefore, pixel-wise background velocity can be calculated by considering the velocity command. As a result, we can obtain the pixel-wise velocity map in xy coordinates by composing human and background velocities of sensor origin. We denote that the x-axis of sensor origin points to the forward direction of the LiDAR.

The parameters for the LiDAR data generation are summarized in Table \ref{table:controlvariable}. These parameters contain depth, xyz coordinates, human label, and velocity map in HDF5 format. We also provide further specific information in the shape of an xml file. Xml files contain a human number in the depth scene, location, weight, and height of each human model. The sampling rates of both human trajectory and LiDAR trajectory are 10 [Hz].

\begin{table*}[t]
	\begin{center}
		\caption{{\bf Comparison of the networks for human segmentation.} Criteria range denotes the evaluation distance range. For example, '0 to 4 [m]' means that the distance range between 0 [m] and 4 [m] from LiDAR was only considered for metric calculation. We would also like to point out that defected pixels in the depth map are not considered for evaluation.}
		{\normalsize
			\begin{tabular}{c|c|c|cccc} \hline
			    & \multirow{2}{*}{ Criteria range} & \multirow{2}{*}{Single frame} & \multicolumn{4}{c}{\footnotesize Sequence}  \\
				 & & & 2-frame & 4-frame & 8-frame & 16-frame \\ \hline % & Ensemble \\ \hline
				\multirow{4}{*}{Generated data}& 0 to $\infty$ [m] & 0.4358 & 0.5869 & 0.5461 & 0.5523 &  0.6443 \\ \cline{2-7}  %& 0.5788 \\
				& 0 to 4 [m] & 0.6301 & 0.7774 & 0.7278 & 0.7250 & 0.8141 \\ % & 0.7910 \\
				& 4 to 8 [m] & 0.2346 & 0.4054 & 0.3740 &  0.3831 & 0.4695 \\ % & 0.4441 \\
				& 8 to $\infty$ [m] & 0.0725 & 0.1323 & 0.1222 & 0.1272 & 0.1622 \\ \hline % & 0.1487 \\ \hline
				\multirow{4}{*}{Real data}& 0 to $\infty$ [m] & 0.5097 & 0.5588 & 0.5531 & 0.5550 & 0.5883 \\ \cline{2-7} % & 0.5651 \\
				& 0 to 4 [m] & 0.6402 & 0.6801 & 0.6837 & 0.6886 & 0.7041 \\ % & 0.6954 \\
				& 4 to 8 [m] &  0.4067 & 0.4543 & 0.4521 & 0.4320 & 0.4668 \\ % & 0.4318 \\
				& 8 to $\infty$ [m] & 0.1883 & 0.2234 & 0.1971 & 0.1965 & 0.2309 \\ \hline% &  0.1950 \\ \hline
			\end{tabular}
		}
		\label{Results_of_networks}
	\end{center}
\end{table*}

\begin{table}[t]
	\begin{center}
		\caption{ Comparison of networks trained with different data }
		{\small
		\begin{tabular}{ccc|c} \hline
			human model & trajectory & walking model & IoU \\ \hline
			\checkmark & &  & 0.2276 \\ 
			\checkmark & \checkmark &  & 0.3318 \\
			\checkmark &  & \checkmark & 0.3395 \\
			\checkmark & \checkmark & \checkmark & {\bf 0.3814} \\ \hline
		\end{tabular}
		}
		\label{exp02}
	\end{center}
\end{table}

\begin{table}[t]
	\begin{center}
		\caption{ Comparison of networks in real data }
		{\normalsize
		\begin{tabular}{c|c} \hline
            Network & IoU \\ \hline
			Single network~\cite{kim2019automatic} & 0.3490 \\ 
			Ensemble network~\cite{kim2019automatic} & 0.3656 \\
            Proposed in 16-frame & {\bf 0.5883} \\ \hline
		\end{tabular}
		}
		\label{networkcomparison}
	\end{center}
\end{table}

\subsection{Training parameters}
\label{policy}
\noindent
We only used generated data for training data in this study. From the Table \ref{table:controlvariable}, the size of the input image is $32 \times 1024$. We employed Adam~\cite{DBLP:journals/corr/KingmaB14} with the learning rate = 0.001 and decay = 0.001 for the optimizer. We also set the weight to human label in every training data as $background$ $pixel$ $number$ / $human$ $pixel$ $number$ for each of the scenes. In addition, we set the weight to categorical cross-entropy loss for segmentation as 10,000, MSE for the background velocity estimation was set to 1, and MSE for human velocity estimation was set to 1,001.

\begin{figure}[t]
	\begin{center}

	\subfloat[Ground truth of generated data in point cloud]{
		\includegraphics[width = 75mm, height = 30mm]{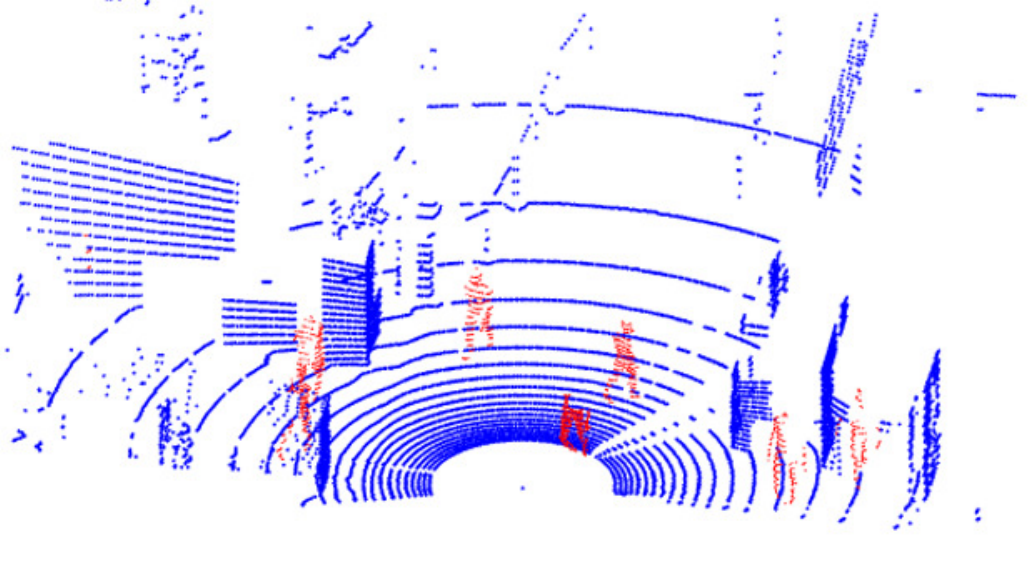}
	} \\
	\subfloat[Prediction with 16-frame in point cloud]{
		\includegraphics[width = 75mm, height = 30mm]{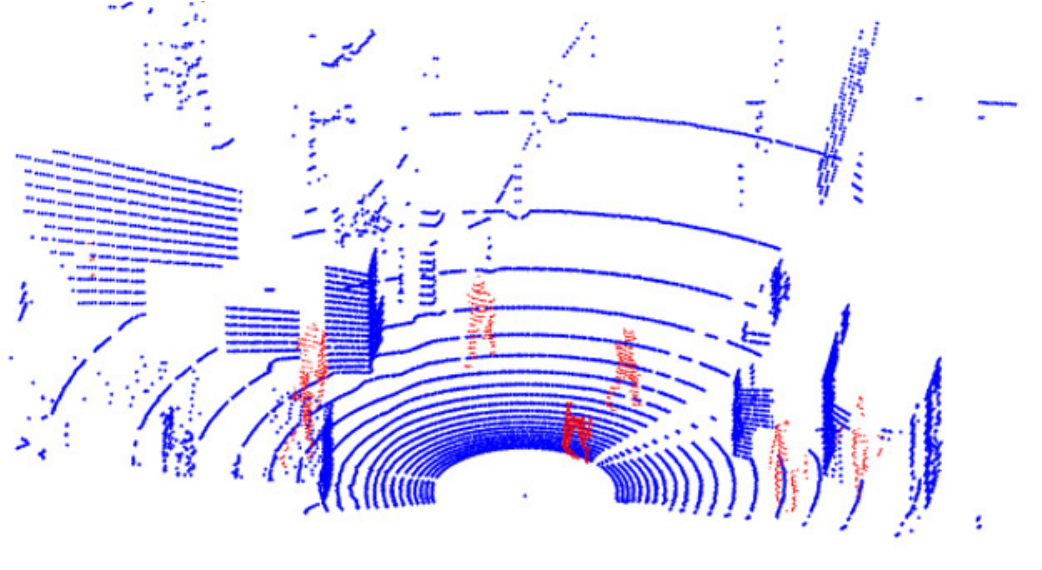}
	} \\
	\caption{{\bfseries An example of segmentation with 16-frame}. In (a), blue points denote the background and red points denote a human label. In (b), blue points denote that the estimated result is the background and red points denote that the estimated result is a human. }
    \label{exampleofprediction_pc}
	\end{center}
\end{figure}

\begin{figure*}[t]
	\begin{flushright}
	
	\subfloat[Color board]{
		\includegraphics[width = 18mm, height = 15mm]{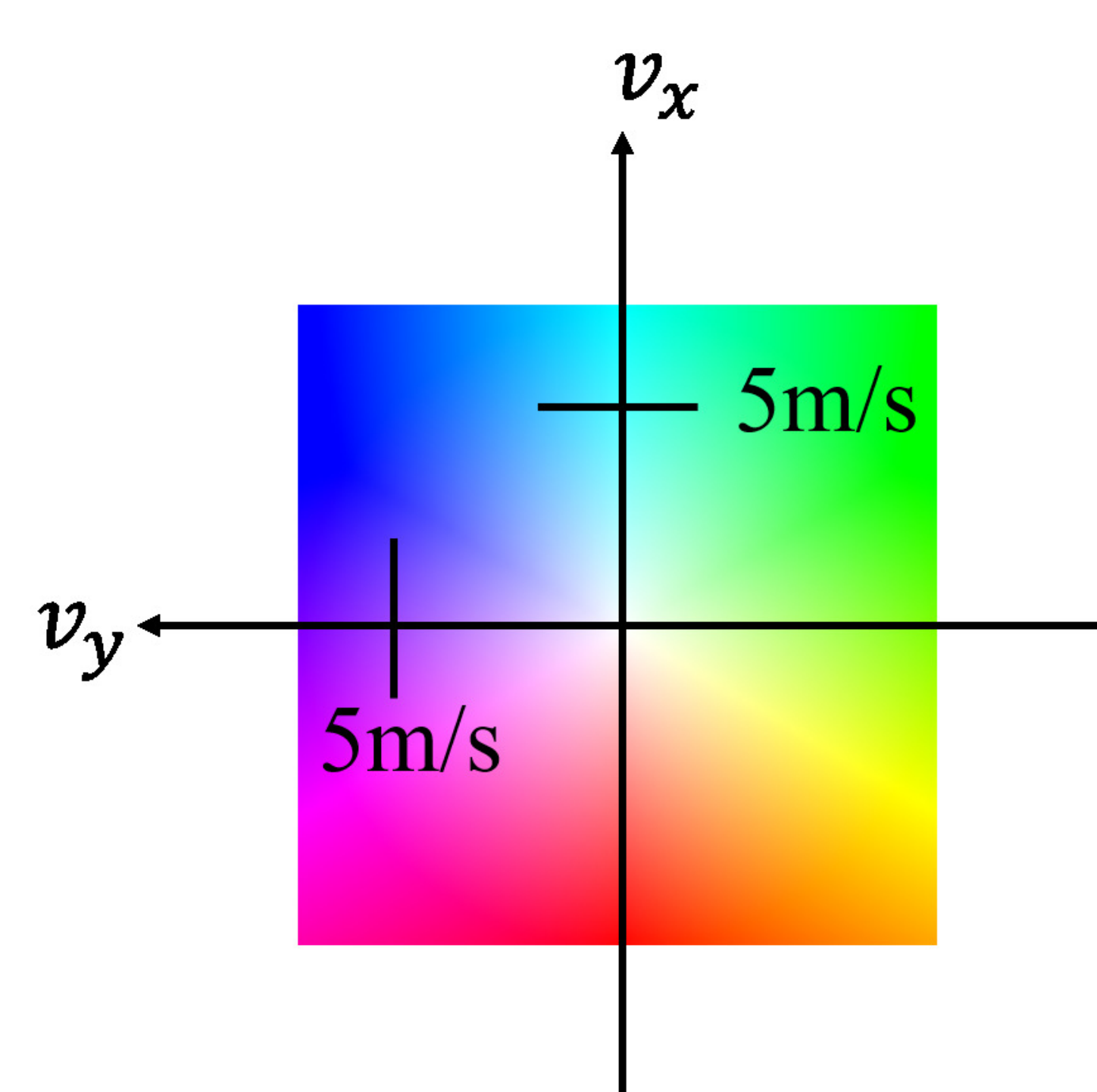}
	}
	\subfloat[Depth image \#01 ]{
		\includegraphics[width = 50mm, height = 10mm]{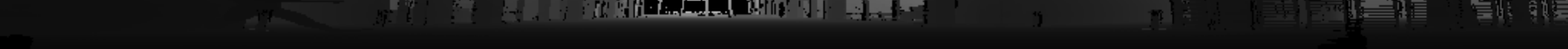}
	}
	\subfloat[Human velocity \#01 ]{
		\frame{\includegraphics[width = 50mm, height = 10mm]{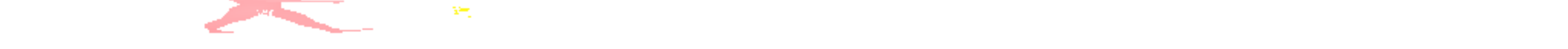}}
	}
	\subfloat[16-frame estimation \#01 ]{
		\frame{\includegraphics[width = 50mm, height = 10mm]{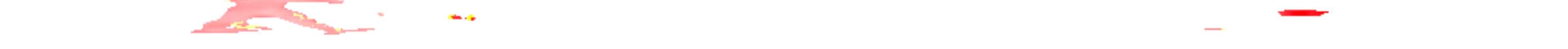}}
	} \\
	\subfloat[Depth image \#02 ]{
		\includegraphics[width = 50mm, height = 10mm]{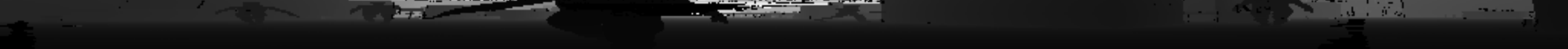}
	}
	\subfloat[Human velocity \#02 ]{
		\frame{\includegraphics[width = 50mm, height = 10mm]{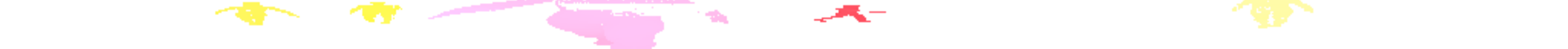}}
	}
	\subfloat[16-frame estimation \#02 ]{
		\frame{\includegraphics[width = 50mm, height = 10mm]{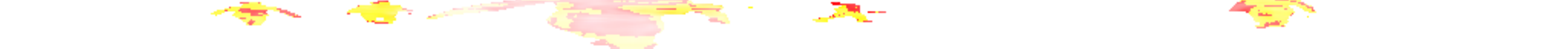}}
	} \\
	\subfloat[Depth image \#03 ]{
		\includegraphics[width = 50mm, height = 10mm]{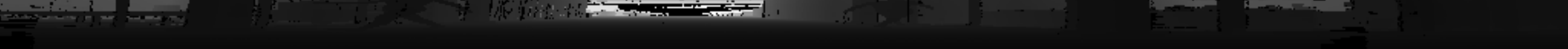}
	}
	\subfloat[Human velocity \#03 ]{
		\frame{\includegraphics[width = 50mm, height = 10mm]{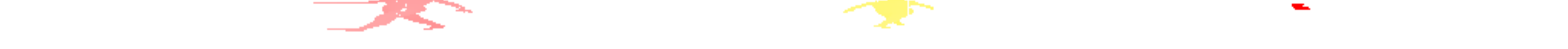}}
	}
	\subfloat[16-frame estimation \#03 ]{
		\frame{\includegraphics[width = 50mm, height = 10mm]{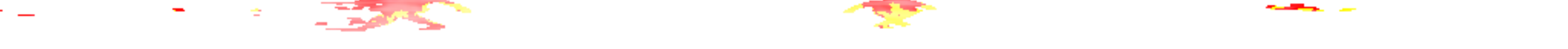}}
	} \\
	\caption{{\bf Estimation of human velocity}. Figures (d), (g), and (j) show the estimation results of human velocity for depth images (b), (e), and (h), respectively, while (c), (f), and (i) show the corresponding ground truth human velocity.
 The meaning of color is shown in Fig. (a).}
   \label{exampleofvelocity}
   \end{flushright}
\end{figure*}

\subsection{Experimental results}
\noindent
We evaluated a single type of network for learning image, and four types of networks for learning sequence. For those networks, we set the parameters as 1,000 for the epoch, 1 for batch size, and 150 for step for epoch. First we trained the single frame network which is a reference of image domain learning with the same architecture in Fig.~\ref{Network architecture for sequence training}. Thus, we set the sampling frame length as 2, 4, 8, and 16 for learning sequence. The manually labeled 0.1K of real data and 108 sequences generated by the proposed pipeline were used for the evaluation. The specific information about real data is described in Table~\ref{realdatatable}. Note that the velocity unit is unified as [m/s] and the training data do not include test data.

\noindent
{\bf Human segmentation.}
Table~\ref{Results_of_networks} shows the evaluation results. From the result, the 'Sequence' shows performance improvement than 'Frame' in every criteria range. In the case of the single frame, it shows comparable performance in '0 to 4 [m]'. However, the performance of the single frame drastically decreases as the criteria range recedes. This decrease problem can be resolved by using a multi frame. For example, there is an improvement of more than 23.4\% in '4 to 8 [m]' and 8.9\% in '8 to $\infty$ [m]' in 16-frame. Furthermore, we can observe that utilizing sequential data also improves human segmentation performance in '0 to 4 [m]' and improves the overall network performance. According to these experimental results, we conclude that utilizing the multi frame can improve performance for recognizing far human than a single frame in a human segmentation task with LiDAR sensing. Inform that the results of 16-frame are used for examples in Fig.~\ref{exampleofprediction_pc} and~\ref{exampleofvelocity} because 16-frame shows the best score in Table~\ref{Results_of_networks}.

To confirm the effect of the sequence feature, we compared four different datasets as shown in Table~\ref{exp02}. In the data that excludes the trajectory and walking model, all human walking models are fixed in a specific point and state. In the data without a walking model, all human walking models are also sliding along their trajectories in a fixed state. In case of the data without trajectory, all human walking models are walking in a fixed position. We trained the 16-frame with 100 sequences in each dataset and evaluated this with real data for the IoU. The results are shown in Table~\ref{exp02}. According to Table~\ref{exp02}, the network trained by data with all information shows the best score. We can also observe that the scores of the network trained by data excluding the walking model and trajectory are higher than data without both trajectory and walking model. Then, we assume that consideration of trajectory and walking model affect performance improvement.

In recent study~\cite{kim2019automatic}, ensemble network was proposed and evaluated using same real data. Therefore, the proposed network was compared with the results reported in~\cite{kim2019automatic} as shown in Table~\ref{networkcomparison}. We can confirm that the proposed network shows siginificantly better score than other networks.  

From the human segmentation results, we assume that the network can extract the difference features in the time axis from sequential data, and the features can improve the segmentation task in far distance . In addition, we also assume that a network within the image domain is capable of learning the human shape only, whereas the video domain is capable of learning both the human shape and its movement. Accordingly, we conclude that the performance decrease problem with distance can be solved by taking time-series information into consideration. The example of human segmentation is illustrated in Fig.~\ref{exampleofprediction_pc}.

\noindent
{\bf Estimation of human velocity.}
As shown in Fig.~\ref{fig:learning_model}, the network can estimate the human segmentation and pixel-wise velocity map. Then, we can derive the velocity map of the segmented area. The examples of human velocity estimation are illustrated in Fig.~\ref{exampleofvelocity}. According to the Fig.~\ref{exampleofvelocity}, the estimated velocities show similar tendency with ground truth. As a consequence, we conclude that velocity estimation with LiDAR only is a feasible task.

\section{Conclusions}
\label{Conclusion}
\noindent
In this paper, we propose a fully automated sequence generation pipeline using a precise human model and motion for human detection with velocity estimation using LiDAR. With this process, we can easily generate labeled data with any properties for LiDAR. Following this result, we conclude that sequential data can improve the performance of human segmentation when compared to data collected from the image domain only. Furthermore, we also confirm the possibility of velocity estimation with LiDAR only. We present 0.1K of labeled real data, and more than 7K of generated sequences with human labels and velocity maps. With these sequences, we were able to confirm the effectiveness of using sequential data over image domain data.

We have considered two main points with regard to future improvements of our work. The first relates to pipeline improvement. Although we used a confirmed method to produce the human model, this is not entirely representative of the real world. As such, we will take into consideration pose, fashion, and other conditions for more accurate simulations. Because we made the human model walk, we will also try to apply different walking poses, stride lengths and other conditions. In addition, it should be noted that networks were only trained with generated data in this study. A comparison of network performance in manually labeled training data and generated training data would be beneficial to this study. However, based on our investigations, this particular experiment is expected to incur significant costs. The second point for improvement relates to utilizing sequences. Because generated data is sequential, this can be applied to many tasks such as human tracking and trajectory prediction. Future work will elaborate on how these points for improvement can be achieved.

\bibliographystyle{IEEEtran}
\bibliography{refs}

\begin{thebibliography}{10}
\providecommand{\url}[1]{#1}
\csname url@rmstyle\endcsname
\providecommand{\newblock}{\relax}
\providecommand{\bibinfo}[2]{#2}
\providecommand\BIBentrySTDinterwordspacing{\spaceskip=0pt\relax}
\providecommand\BIBentryALTinterwordstretchfactor{4}
\providecommand\BIBentryALTinterwordspacing{\spaceskip=\fontdimen2\font plus
\BIBentryALTinterwordstretchfactor\fontdimen3\font minus
  \fontdimen4\font\relax}
\providecommand\BIBforeignlanguage[2]{{%
\expandafter\ifx\csname l@#1\endcsname\relax
\typeout{** WARNING: IEEEtran.bst: No hyphenation pattern has been}%
\typeout{** loaded for the language `#1'. Using the pattern for}%
\typeout{** the default language instead.}%
\else
\language=\csname l@#1\endcsname
\fi
#2}}

\bibitem{schwarz2010lidar}
B.~Schwarz, ``Lidar: Mapping the world in 3d,'' \emph{Nature Photonics},
  vol.~4, no.~7, p. 429, 2010.

\bibitem{niijima2018autonomous}
S.~Niijima, Y.~Sasaki, and H.~Mizoguchi, ``Autonomous navigation of electric
  wheelchairs in urban areas on the basis of self-generated 2d drivable maps,''
  in \emph{2018 IEEE/ASME International Conference on Advanced Intelligent
  Mechatronics (AIM)}.\hskip 1em plus 0.5em minus 0.4em\relax IEEE, 2018, pp.
  1081--1086.

\bibitem{jiang2014thumos}
Y.~G. Jiang, J.~Liu, A.~R. Zamir, G.~Toderici, I.~Laptev, M.~Shah, and
  R.~Sukthankar, ``Thumos challenge: Action recognition with a large number of
  classes,'' 2014.

\bibitem{Li_2015_ICCV}
Y.~Li, X.~Yang, and J.~Luo, ``Semantic video entity linking based on visual
  content and metadata,'' in \emph{The IEEE International Conference on
  Computer Vision (ICCV)}, December 2015.

\bibitem{boiman2007detecting}
O.~Boiman and M.~Irani, ``Detecting irregularities in images and in video,''
  \emph{International journal of computer vision}, vol.~74, no.~1, pp. 17--31,
  2007.

\bibitem{Jonathan}
J.~Long, S.~Evan, and D.~Trevor, ``Fully convolutional networks for semantic
  segmentation.'' \emph{Proceedings of the IEEE conference on computer vision
  and pattern recognition (CVPR)}, pp. 3431--3440, 2015.

\bibitem{Farabet}
C.~Farabet, C.~Camille, N.~Laurent, and L.~Yann, ``Learning hierarchical
  features for scene labeling.'' \emph{IEEE transactions on pattern analysis
  and machine intelligence}, vol.~35, pp. 1915--1929, 2013.

\bibitem{DBLP}
\BIBentryALTinterwordspacing
B.~Wu, A.~Wan, X.~Yue, and K.~Keutzer, ``Squeezeseg: Convolutional neural nets
  with recurrent {CRF} for real-time road-object segmentation from 3d lidar
  point cloud,'' \emph{CoRR}, vol. abs/1710.07368, 2017. [Online]. Available:
  \url{http://arxiv.org/abs/1710.07368}
\BIBentrySTDinterwordspacing

\bibitem{SlowFast}
\BIBentryALTinterwordspacing
C.~Feichtenhofer, H.~Fan, J.~Malik, and K.~He, ``Slowfast networks for video
  recognition,'' \emph{CoRR}, vol. abs/1812.03982, 2018. [Online]. Available:
  \url{http://arxiv.org/abs/1812.03982}
\BIBentrySTDinterwordspacing

\bibitem{Tran_2015_ICCV}
D.~Tran, L.~Bourdev, R.~Fergus, L.~Torresani, and M.~Paluri, ``Learning
  spatiotemporal features with 3d convolutional networks,'' in \emph{The IEEE
  International Conference on Computer Vision (ICCV)}, December 2015.

\bibitem{Gu_2018_CVPR}
C.~Gu, C.~Sun, D.~A. Ross, C.~Vondrick, C.~Pantofaru, Y.~Li,
  S.~Vijayanarasimhan, G.~Toderici, S.~Ricco, R.~Sukthankar, C.~Schmid, and
  J.~Malik, ``Ava: A video dataset of spatio-temporally localized atomic visual
  actions,'' in \emph{The IEEE Conference on Computer Vision and Pattern
  Recognition (CVPR)}, June 2018.

\bibitem{Hong_2017_CVPR}
S.~Hong, D.~Yeo, S.~Kwak, H.~Lee, and B.~Han, ``Weakly supervised semantic
  segmentation using web-crawled videos,'' in \emph{The IEEE Conference on
  Computer Vision and Pattern Recognition (CVPR)}, July 2017.

\bibitem{Seguin_2016_CVPR}
G.~Seguin, P.~Bojanowski, R.~Lajugie, and I.~Laptev, ``Instance-level video
  segmentation from object tracks,'' in \emph{The IEEE Conference on Computer
  Vision and Pattern Recognition (CVPR)}, June 2016.

\bibitem{wkim}
W.~Kim, M.~Tanaka, M.~Okutomi, and Y.~Sasaki, ``Automatic labeled lidar data
  generation based on precise human model,'' \emph{International Conference on
  Robotics and Automation}, 2019.

\bibitem{kim2019automatic}
------, ``Automatic labeled lidar data generation and distance-based ensemble
  learning for human segmentation,'' \emph{IEEE Access}, vol.~7, pp.
  55\,132--55\,141, 2019.

\bibitem{mochimaru2006dhaiba}
M.~Mochimaru, M.~Kouchi, N.~Miyata, M.~Tada, Y.~Yoshida, K.~Aoki, K.~Kawachi,
  and Y.~Motomura, ``Dhaiba: Functional human models to represent variation of
  shape, motion and subjective assessment,'' SAE Technical Paper, Tech. Rep.,
  2006.

\bibitem{Kinect}
Microsoft,
  https://news.microsoft.com/2010/11/04/the-future-of-entertainment-starts-today-as-kinect-for-xbox-360-leaps-and-lands-at-retailers-nationwide/,
  accessed in 2019 Jul 17.

\bibitem{barbosa2012re}
I.~B. Barbosa, M.~Cristani, A.~Del~Bue, L.~Bazzani, and V.~Murino,
  ``Re-identification with rgb-d sensors,'' in \emph{European Conference on
  Computer Vision}.\hskip 1em plus 0.5em minus 0.4em\relax Springer, 2012, pp.
  433--442.

\bibitem{munaro20143d}
M.~Munaro, A.~Basso, A.~Fossati, L.~Van~Gool, and E.~Menegatti, ``3d
  reconstruction of freely moving persons for re-identification with a depth
  sensor,'' in \emph{Robotics and Automation (ICRA), 2014 IEEE International
  Conference on}.\hskip 1em plus 0.5em minus 0.4em\relax IEEE, 2014, pp.
  4512--4519.

\bibitem{kastaniotis2015framework}
D.~Kastaniotis, I.~Theodorakopoulos, C.~Theoharatos, G.~Economou, and
  S.~Fotopoulos, ``A framework for gait-based recognition using kinect,''
  \emph{Pattern Recognition Letters}, vol.~68, pp. 327--335, 2015.

\bibitem{Geiger2012CVPR}
A.~Geiger, P.~Lenz, and R.~Urtasun, ``Are we ready for autonomous driving? the
  kitti vision benchmark suite,'' in \emph{Conference on Computer Vision and
  Pattern Recognition (CVPR)}, 2012.

\bibitem{premebida2014pedestrian}
C.~Premebida, J.~Carreira, J.~Batista, and U.~Nunes, ``Pedestrian detection
  combining rgb and dense lidar data,'' in \emph{Intelligent Robots and Systems
  (IROS 2014), 2014 IEEE/RSJ International Conference on}.\hskip 1em plus 0.5em
  minus 0.4em\relax IEEE, 2014, pp. 4112--4117.

\bibitem{levi2015stixelnet}
D.~Levi, N.~Garnett, E.~Fetaya, and I.~Herzlyia, ``Stixelnet: A deep
  convolutional network for obstacle detection and road segmentation.'' in
  \emph{BMVC}, 2015, pp. 109--1.

\bibitem{johnson2017driving}
M.~Johnson-Roberson, C.~Barto, R.~Mehta, S.~N. Sridhar, K.~Rosaen, and
  R.~Vasudevan, ``Driving in the matrix: Can virtual worlds replace
  human-generated annotations for real world tasks?'' in \emph{Robotics and
  Automation (ICRA), 2017 IEEE International Conference on}.\hskip 1em plus
  0.5em minus 0.4em\relax IEEE, 2017, pp. 746--753.

\bibitem{richter2016playing}
S.~R. Richter, V.~Vineet, S.~Roth, and V.~Koltun, ``Playing for data: Ground
  truth from computer games,'' in \emph{European Conference on Computer
  Vision}.\hskip 1em plus 0.5em minus 0.4em\relax Springer, 2016, pp. 102--118.

\bibitem{brox2010object}
T.~Brox and J.~Malik, ``Object segmentation by long term analysis of point
  trajectories,'' in \emph{European conference on computer vision}.\hskip 1em
  plus 0.5em minus 0.4em\relax Springer, 2010, pp. 282--295.

\bibitem{Tsai2010MotionCT}
D.~Tsai, M.~Flagg, and J.~M. Rehg, ``Motion coherent tracking with multi-label
  mrf optimization,'' in \emph{BMVC}, 2010.

\bibitem{li2013video}
F.~Li, T.~Kim, A.~Humayun, D.~Tsai, and J.~M. Rehg, ``Video segmentation by
  tracking many figure-ground segments,'' in \emph{Proceedings of the IEEE
  International Conference on Computer Vision}, 2013, pp. 2192--2199.

\bibitem{caelles20182018}
S.~Caelles, A.~Montes, K.-K. Maninis, Y.~Chen, L.~Van~Gool, F.~Perazzi, and
  J.~Pont-Tuset, ``The 2018 davis challenge on video object segmentation,''
  \emph{arXiv preprint arXiv:1803.00557}, vol.~1, no.~2, 2018.

\bibitem{APOLLO}
APOLLO, http://apollo.auto/index.html, accessed in 2019 Jul 27.

\bibitem{Geiger2013IJRR}
A.~Geiger, P.~Lenz, C.~Stiller, and R.~Urtasun, ``Vision meets robotics: The
  kitti dataset,'' \emph{International Journal of Robotics Research (IJRR)},
  2013.

\bibitem{Cordts2016Cityscapes}
M.~Cordts, M.~Omran, S.~Ramos, T.~Rehfeld, M.~Enzweiler, R.~Benenson,
  U.~Franke, S.~Roth, and B.~Schiele, ``The cityscapes dataset for semantic
  urban scene understanding,'' in \emph{Proc. of the IEEE Conference on
  Computer Vision and Pattern Recognition (CVPR)}, 2016.

\bibitem{endo2014hand}
Y.~Endo, M.~Tada, and M.~Mochimaru, ``Hand modeling and motion reconstruction
  for individuals,'' \emph{Int. J. of Automation Technology Vol}, vol.~8,
  no.~3, 2014.

\bibitem{endo2015estimation}
------, ``Estimation of arbitrary human models from anthropometric
  dimensions,'' in \emph{International Conference on Digital Human Modeling and
  Applications in Health, Safety, Ergonomics and Risk Management}.\hskip 1em
  plus 0.5em minus 0.4em\relax Springer, 2015, pp. 3--14.

\bibitem{Mirai}
Miraikan, https://www.miraikan.jst.go.jp, accessed in 2018 May 27.

\bibitem{threatscore}
I.~T. Jolliffe and D.~B. Stephenson, \emph{Forecast verification: a
  practitioner's guide in atmospheric science}.\hskip 1em plus 0.5em minus
  0.4em\relax John Wiley \& Sons, 2003.

\bibitem{ohsato2015real}
A.~Ohsato, Y.~Sasaki, and H.~Mizoguchi, ``Real-time 6dof localization for a
  mobile robot using pre-computed 3d laser likelihood field,'' in
  \emph{Robotics and Biomimetics (ROBIO), 2015 IEEE International Conference
  on}.\hskip 1em plus 0.5em minus 0.4em\relax IEEE, 2015, pp. 2359--2364.

\bibitem{kobayashi2016age}
Y.~Kobayashi, H.~Hobara, T.~A. Heldoorn, M.~Kouchi, and M.~Mochimaru,
  ``Age-independent and age-dependent sex differences in gait pattern
  determined by principal component analysis,'' \emph{Gait \& posture},
  vol.~46, pp. 11--17, 2016.

\bibitem{Hokuyo}
HOKUYO, https://www.hokuyo-aut.jp/search/single.php?serial=169, accessed in
  2019 Feb 19.

\bibitem{sasaki2017long}
Y.~Sasaki and J.~Nitta, ``Long-term demonstration experiment of autonomous
  mobile robot in a science museum,'' in \emph{2017 IEEE International
  Symposium on Robotics and Intelligent Sensors (IRIS)}.\hskip 1em plus 0.5em
  minus 0.4em\relax IEEE, 2017, pp. 304--310.

\bibitem{DBLP:journals/corr/KingmaB14}
\BIBentryALTinterwordspacing
D.~P. Kingma and J.~Ba, ``Adam: {A} method for stochastic optimization,''
  \emph{CoRR}, vol. abs/1412.6980, 2014. [Online]. Available:
  \url{http://arxiv.org/abs/1412.6980}
\BIBentrySTDinterwordspacing

\end{thebibliography}
\EOD
\end{document}